\documentclass[11pt,a4paper]{article}
\usepackage{geometry}
\usepackage{authblk}
\usepackage[hyperref]{emnlp2020}
\usepackage{times}
\usepackage{latexsym}

\usepackage{times}
\usepackage{latexsym}
\usepackage{amsmath}
\usepackage{url}
\usepackage{mathdots}
\usepackage{yhmath}
\usepackage{cancel}
\usepackage{color}
\usepackage{siunitx}
\usepackage{array}
\usepackage{multirow}
\usepackage{amssymb}
\usepackage{gensymb}
\usepackage{graphicx}
\usepackage{float}
\usepackage{rotating}
\usepackage{tabularx}
\usepackage{booktabs}
\usepackage{svg}
\usepackage{tablefootnote}
\usepackage{cleveref}
\usepackage{tabularx}
\usepackage{xcolor}
\usepackage{makecell}
\usepackage{enumitem}
\usepackage{multirow}

\usepackage{microtype}

\aclfinalcopy 

\newtoggle{showtodos}
\toggletrue{showtodos}

\newcommand{\cut}[1]{}
\newcommand{\personachat}{\textsc{Persona-Chat}}
\newcommand{\compac}{{\textsc{Compac}}}

\newcommand{\newpara}[1]{\noindent \textbf{#1} \hspace{0.5em}}

\makeatletter
\renewcommand*{\@fnsymbol}[1]{\ensuremath{\ifcase#1\or \dagger\or \ddagger\or
    \mathsection\or \mathparagraph\or \|\or **\or \dagger\dagger
    \or \ddagger\ddagger \else\@ctrerr\fi}}
\makeatother

\title{Like hiking? You probably \textit{enjoy nature}:\\ Persona-grounded Dialog with Commonsense Expansions}

\author[$\clubsuit$]{\textbf{Bodhisattwa Prasad Majumder}}
\author[$\spadesuit$]{\textbf{Harsh Jhamtani}}
\author[$\clubsuit$]{\textbf{\qquad \qquad \qquad \qquad \qquad Taylor Berg-kirkpatrick}}
\author[$\clubsuit$]{\textbf{Julian McAuley}}
\affil[$\clubsuit$]{Department of Computer Science and Engineering, UC San Diego \protect\\ \tt \{bmajumde, tberg, jmcauley\}@eng.ucsd.edu}
\affil[$\spadesuit$]{School of Computer Science, Carnegie Mellon University \protect\\ \tt jharsh@cs.cmu.edu}

\makeatletter
\renewcommand\outauthor{
    \begin{tabular}[t]{>{\centering}p{14cm}} 
    \bf\@author
    \fi
    \end{tabular}}
\makeatother

\date{}

\begin{document}
\maketitle
\begin{abstract}

Existing persona-grounded dialog models often fail to capture simple implications of given persona descriptions, something which humans 
are able to do
seamlessly.
For example, state-of-the-art models cannot infer that
interest in hiking might imply love for nature or longing for a break.
In this paper, we propose to expand available persona sentences using existing commonsense knowledge bases and paraphrasing resources 
to imbue
dialog models with access to an expanded and richer set of persona
descriptions.
Additionally, we introduce \textit{fine-grained} grounding on personas by encouraging the model to make a discrete choice among persona sentences while synthesizing a dialog response.
Since such a choice is not observed in the data, we model it using a discrete latent random variable and
use variational learning
to sample from hundreds of persona expansions.
Our model outperforms competitive baselines on the \personachat~dataset in terms of dialog quality and diversity while achieving persona-consistent and controllable dialog generation.

\end{abstract}

\section{Introduction}

\begin{figure}[t!]
    \centering
    \includegraphics[width=0.9\linewidth]{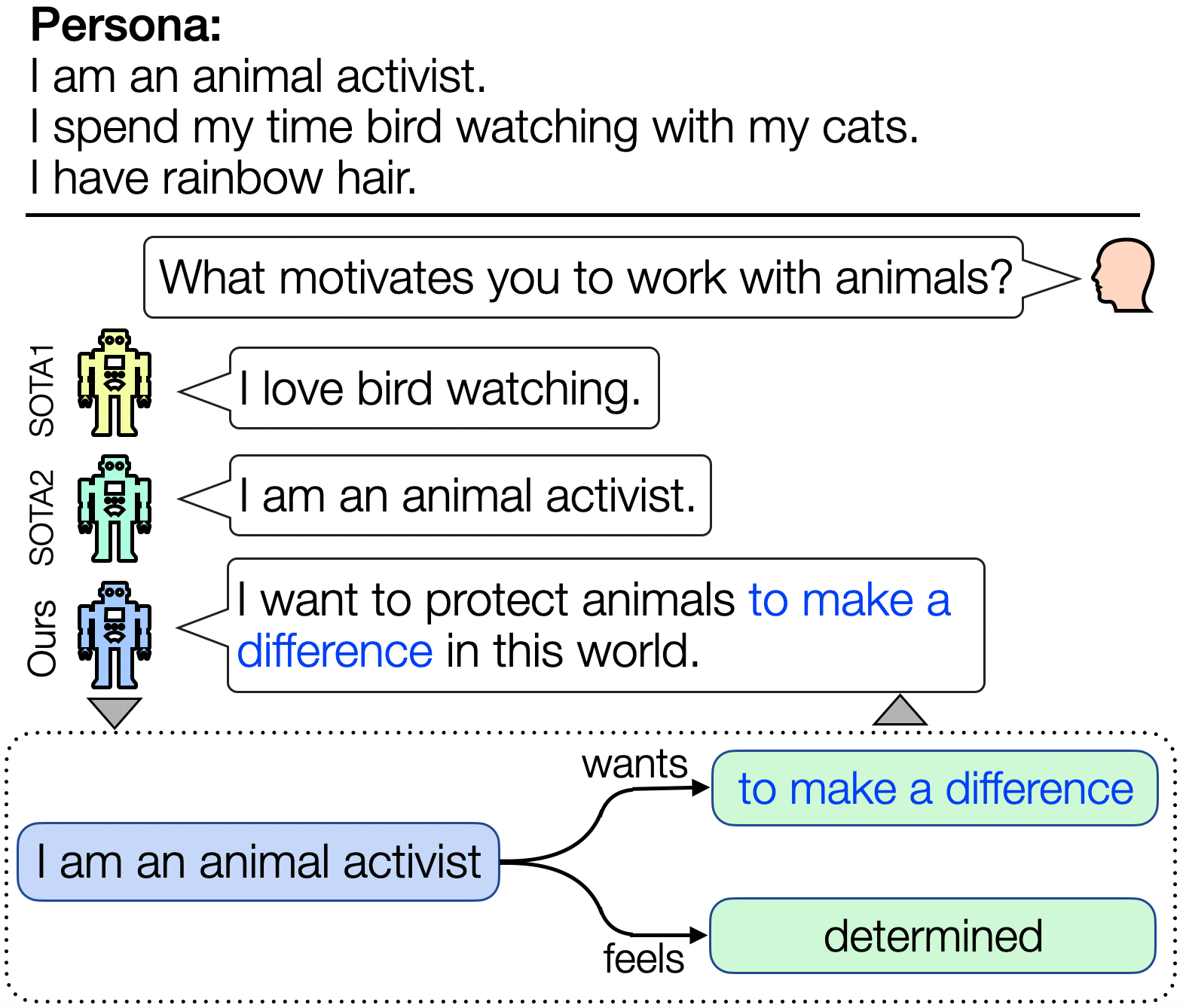}
    \caption{State-of-the-art models struggle to respond a user's query,
    where generating an engaging response depends on commonsense reasoning.
    }
    \label{fig:fig1}
\end{figure}

Persona-grounded dialog generation is a `chit-chat' dialog setup where a dialog agent is expected to communicate based on a given profile \cite{DBLP:conf/acl/KielaWZDUS18}. Many recent works have focused on a popular benchmark dataset for this task: \personachat~\cite{DBLP:conf/acl/KielaWZDUS18} that provides personas as a set of sentences along with each dialog (example in \Cref{fig:fig1}). However, a careful analysis of state-of-the-art (SOTA) models reveals that they often struggle to respond to contexts that do not closely match given persona sentences,
even when the implications might be obvious to a human.

For example, in \Cref{fig:fig1}, the user asks an \textit{indirect} question to the bot related to one of its persona sentences: 
\textit{I am an animal activist}.
SOTA1, which concatenates all persona sentences with dialog history and finetunes a pre-trained generative model (e.g.~GPT2) \cite{DBLP:journals/corr/abs-1901-08149}, fails to infer implied commonsense from the dialog context and conditions on an incorrect persona. 
SOTA2, which separately selects a persona sentence given the dialog history \cite{DBLP:conf/ijcai/LianXWPW19} manages to choose the correct persona but merely copies it as the final response.
Neither approach is in general
capable of responding to context that goes beyond what is explicitly mentioned in the available persona sentences, 
which limits consistent and interesting conversation.
The goal of our model is to
understand that being `an animal activist' may imply that the person wants `to make a difference' via their activity towards animals and synthesizes a context-consistent and engaging response.

\begin{figure*}[t!]
\includegraphics[width=0.8\linewidth]{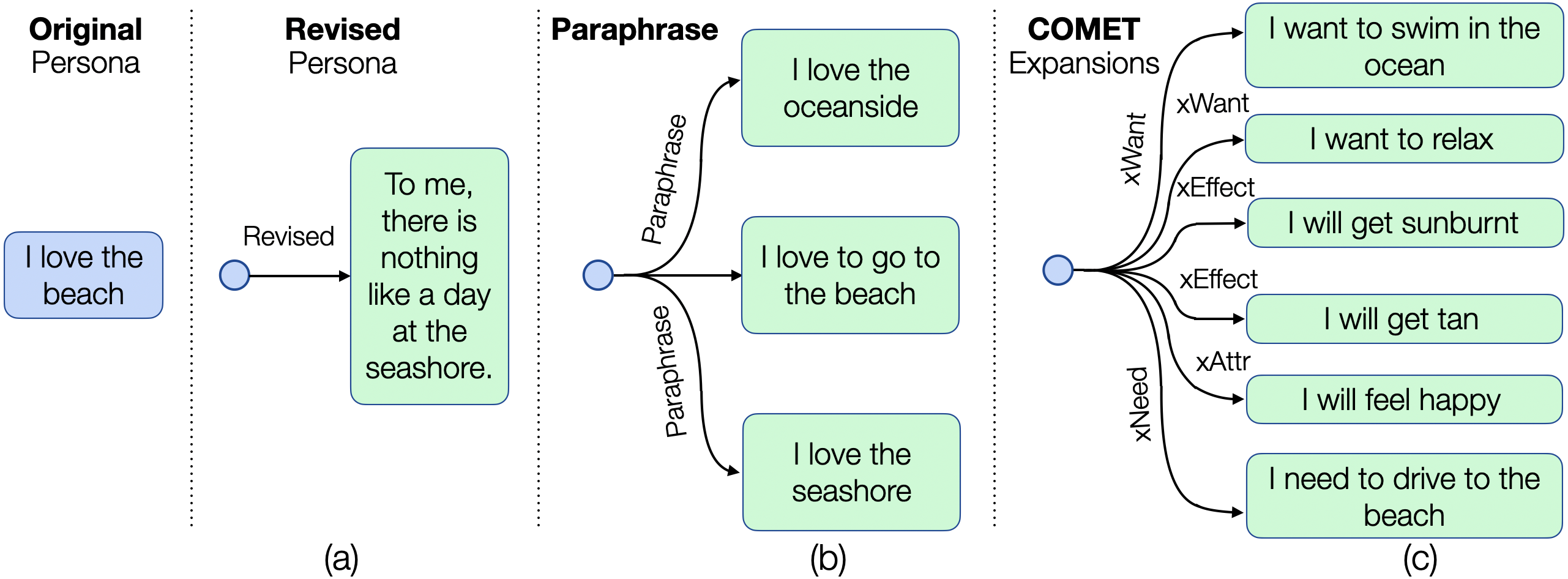}
\centering
\caption{Expansions of an original persona via (a) human rewrite \shortcite{DBLP:conf/acl/KielaWZDUS18}, (b) paraphrase, and (c) COMET.
}
\label{fig:expansion}
\end{figure*}

In this paper, we focus on making persona-grounded chatbots more consistent with personas and implicit dialog context.
We present a framework to expand available persona sentences to their commonsense implications by using an existing commonsense knowledge base or paraphrasing resources (see \Cref{sec:expansion}).
We endow our dialog model with these expansions directly 
rather than 
requiring the model to
learn them from scratch for being context-consistent.
We 
find
that expansions derived from a commonsense knowledge base are more useful to provide engaging contextual information compared to other expansion sources.

We further propose a \textbf{Com}mon Sense and \textbf{P}ersona \textbf{A}ligned \textbf{C}hatbot\footnote{Code is available at -- \url{https://github.com/majumderb/compac}.} (\textbf{\compac})
which models choices over the \textit{expanded} persona set via a discrete latent random variable (See \Cref{sec:method}) as \textit{fine-grained} persona grounding.
Even though it is tractable to marginalize over all expansions, that would require a forward pass through the dialog generator for each outcome which is prohibitively slow during training.
Instead, to accommodate hundreds of persona expansions, we train the model by optimizing a lower bound on the log-likelihood.
We use amortized variational inference by approximating the true posterior using an inference network that eventually provides useful inductive bias.
Particularly, we show that our Bayesian formulation for the fine-grained persona grounding was essential as simply providing expanded knowledge does not help the model generate better responses.

We also outperform competitive baselines in all dialog quality metrics as well as human evaluations which find \compac~to be engaging and coherent.
We demonstrate that \compac~learns to be consistent with the dialog context with accurate persona grounding especially in the presence of commonsense expansions. Finally, we show that our model can reflect a change in response generation when a grounding persona is modified, indicating the possibility of controllable generation.
\section{Persona Grounded Dialog}
We 
use
a popular benchmark dataset: \personachat~\cite{DBLP:conf/acl/KielaWZDUS18} for our persona-grounded dialog generation task.
It contains 10,907 dialogs between pairs of speakers where each speaker follows their own persona; 968 dialogs are used for validation and 1,000 for testing.
Each speaker is described by 3-5 persona sentences. (e.g.~`I love the beach', `My mother is a medical doctor').
Out of 1,155 total unique personas, 100 are used for validation and 100 for testing.

The task of persona-grounded dialog generation is: given a dialog history $H$ and grounding persona sentences $S$, we must predict the next utterance $x$ (Summary of notations in \Cref{tab:notations}).
Hence a dialog model should maximize the likelihood $p(x|H, S)$.  
From
the
\personachat~dataset, we use 131,438 utterances for training the dialog model, 15,602 for validation, and 15,024 for testing.


\section{Persona Expansion \cut{via Structured commonsense Knowledge Base}}
\label{sec:expansion}
Persona sentences used in persona-grounded dialogs are instances of world events that often imply real-world consequences or richer information.
For example, `I love surfing' naturally implies that the person might be `adventurous' or `loves the outdoors'.
Similarly, it also means that the person
wants `to go to the beach' regularly.
Inferring these \textit{expansions} from the original fact is non-trivial without additional commonsense knowledge.

\citet{DBLP:conf/acl/KielaWZDUS18} found evidence that having human  written interpretations of a persona sentence via rephrasing
often helps in providing novel information in persona grounding.
While obtaining such expansions by manual rewriting is expensive,
here we explore two 
automatic ways to generate them at scale and separately evaluate them on the downstream dialog modeling task.


\subsection{COMET}
\label{sec:comet}
COMET \cite{DBLP:conf/acl/BosselutRSMCC19} is a framework that generates rich and diverse commonsense expansions of a given world event.
It is a finetuned version of a pre-trained GPT2 \cite{Radford2018ImprovingLU} model on a pre-existing commonsense knowledge graph such as ATOMIC \cite{DBLP:conf/aaai/SapBABLRRSC19} that can generate novel nodes (events) and edges (relations), as seen in \Cref{fig:expansion}c. 
Specifically,
ATOMIC provides tuples that belong to nine relation types spanning over cause-effect interrelations between events: \texttt{oEffect}, \texttt{oReact}, \texttt{oWant}, \texttt{xAttr}, \texttt{xEffect}, \texttt{xIntent}, \texttt{xNeed}, \texttt{xReact}, and \texttt{xWant}---where a prefix `\texttt{x}' indicates an effect or cause on the person
and `\texttt{o}' denotes the same on others.
While we tried COMET finetuned on an alternative commonsense knowledge base (e.g.)~ConceptNet, not all of the expansions were appropriate to describe a persona, mainly because we observe that persona sentences are \textit{event}-like (`I love to go to the beach') as opposed to \textit{concepts} such as `beach'.
For more details on COMET and ATOMIC we refer
the reader 
to \cite{DBLP:conf/acl/BosselutRSMCC19} and \cite{DBLP:conf/aaai/SapBABLRRSC19} respectively.

We use the COMET framework to generate expansions for each persona sentence along the 
nine relation types that ATOMIC provides.
We obtain different samples while decoding via beam search from COMET for more diverse and unique expansions, as shown in \Cref{fig:expansion}c.
We preprocess these expansions to add suitable prefixes to make them similar to the original persona. For example, expansions relating to \texttt{xWant} and \texttt{xAttr} are prefixed with `I want' and `I am' respectively.
For each persona sentence, we generate 5 expansions per relation, i.e.,~in total we will obtain $5\times9 = 45$ expansions per persona sentence.



\subsection{Paraphrasing}

To explore alternative sources for generating commonsense expansions beyond COMET, we consider paraphrasing persona sentences.
Paraphrases of a sentence convey almost the same meaning to a listener as the original. Often paraphrases use synonymous phrases or manipulate word-syntax of the original sentence, which implicitly involves both context comprehension and world knowledge \cite{DBLP:journals/access/ZengZXWJ19}. We obtain these in two ways:

\newpara{Paraphrase Network}
To generate paraphrases at scale, we use an off-the-shelf paraphrasing system based on back-translation \cite{DBLP:journals/corr/abs-1904-12848, DBLP:conf/aclnut/FedermannEQ19} with
pre-trained language translation models.
We make use of \texttt{En-Fr} and \texttt{Fr-En} pre-trained translation models as the components for back-translation.\footnote{\url{https://github.com/google-research/uda}}
While we tried 
other language pairs,
the
\texttt{En-Fr} pair 
proved the most satisfactory
based on qualitative analysis on 500 samples.
We generate 5 paraphrases per persona sentence, 
which
readily provides 
more lexical and syntactic variants
as shown in \Cref{fig:expansion}b.

\newpara{Manual Paraphrasing}
To compare with other expansions, we reuse manually written revised versions of persona sentences provided with \personachat~\cite{DBLP:conf/acl/KielaWZDUS18} though these are limited to only one paraphrase per sentence. We call them \textbf{revised} for short (see \Cref{fig:expansion}a).


\section{Common sense and Persona Aligned Chatbot (\compac)}
\label{sec:method}

To infuse commonsense context in persona-grounded dialog generation, we imbue our dialog model with the expanded persona set instead of only original personas $S$.
But these persona expansions lead to hundreds of new sentences as opposed to only a few given persona sentences which makes it infeasible to encode using a single transformer, as was done in prior works \cite{DBLP:journals/corr/abs-1901-08149}. Additionally, encoding all persona sentences as a single text input leads to a lack of interpretability i.e.,~it is not clear which persona
sentence
was used by the model in generating a particular response.

Instead, we propose \textbf{\compac}:~\textbf{Com}mon Sense and \textbf{P}ersona \textbf{A}ligned \textbf{C}hatbot that allows us to make a \textit{fine-grained} choice of a persona sentence to generate the target response. 
Let $C$ denote a list of expended personas, derived from $S$
(including $S$ itself).
We further add a null persona $\varnothing$ in $C$ considering that some utterances can purely condition on the dialog context.
We are interested in modeling the conditional $p(x|H, C) = p(z|H,C) p(x|z,H,C)$ where $z \in \lbrace 1,2,\ldots,|C| \rbrace$ is a latent discrete random variable, unobserved in the data.
Given the dialog history $H$, first we sample a particular persona sentence $C_{z}$ from a \textit{prior network} $p_{\theta}(z|H)$ (see \Cref{fig:model}).
Next, as depicted in \Cref{fig:model}, the dialog response $x$ is sampled from a \textit{generator network} $p_{\phi}(x|H, C_z)$ by conditioning on the history $H$ and chosen persona sentence $C_z$.

In the generative model described above, the latent variable $z$ is a discrete random variable which points to a single persona sentence. This decision (of conditioning on a single persona sentence) was based on the observation that most dialog responses in the datasets under consideration are relevant to only one persona sentence. It is possible to allow for multiple persona sentences by defining $z$ to pick a subset of $|C|$ persona sentences instead of picking a single sentence. We leave this as a possible future extension.

\begin{table}[t!]
\small
\centering
\begin{tabular}{@{}ll@{}}
\toprule
$S$ & Set of original persona sentences \\
$C$ & \begin{tabular}{@{}l@{}}Set of expanded persona sentences (includes $S$\\and a null persona $\varnothing$) \end{tabular} \\
$H$ & Dialog history with alternative turns from each speaker \\
$x$ & Target utterance \\
$z$ & Discrete latent random variable $\in \lbrace 1,2,\ldots,|C| \rbrace$ \\
$e$ & Mean of RoBERTa subword embeddings as an encoder \\
$t_k$ & Expansion type for $k$-th expansion \\
$f_i$ & $i$-th feature function for prior network; $i \in \lbrace 1,2,3 \rbrace$ \\
$\theta$ & Parameters for prior network $p_{\theta}(z| H, C)$ \\
$\phi$ & Parameters for generator network $p_{\phi}(x| H, C_z)$ \\
$\alpha$ & Parameters for inference network $p_{\alpha}(z|x, H, C)$
\\ \bottomrule
\end{tabular}
\caption{Summary of notation used in the paper}
\label{tab:notations}
\end{table}

\subsection{Persona Choice Prior}
\begin{figure}[t]
\includegraphics[width=0.95\linewidth]{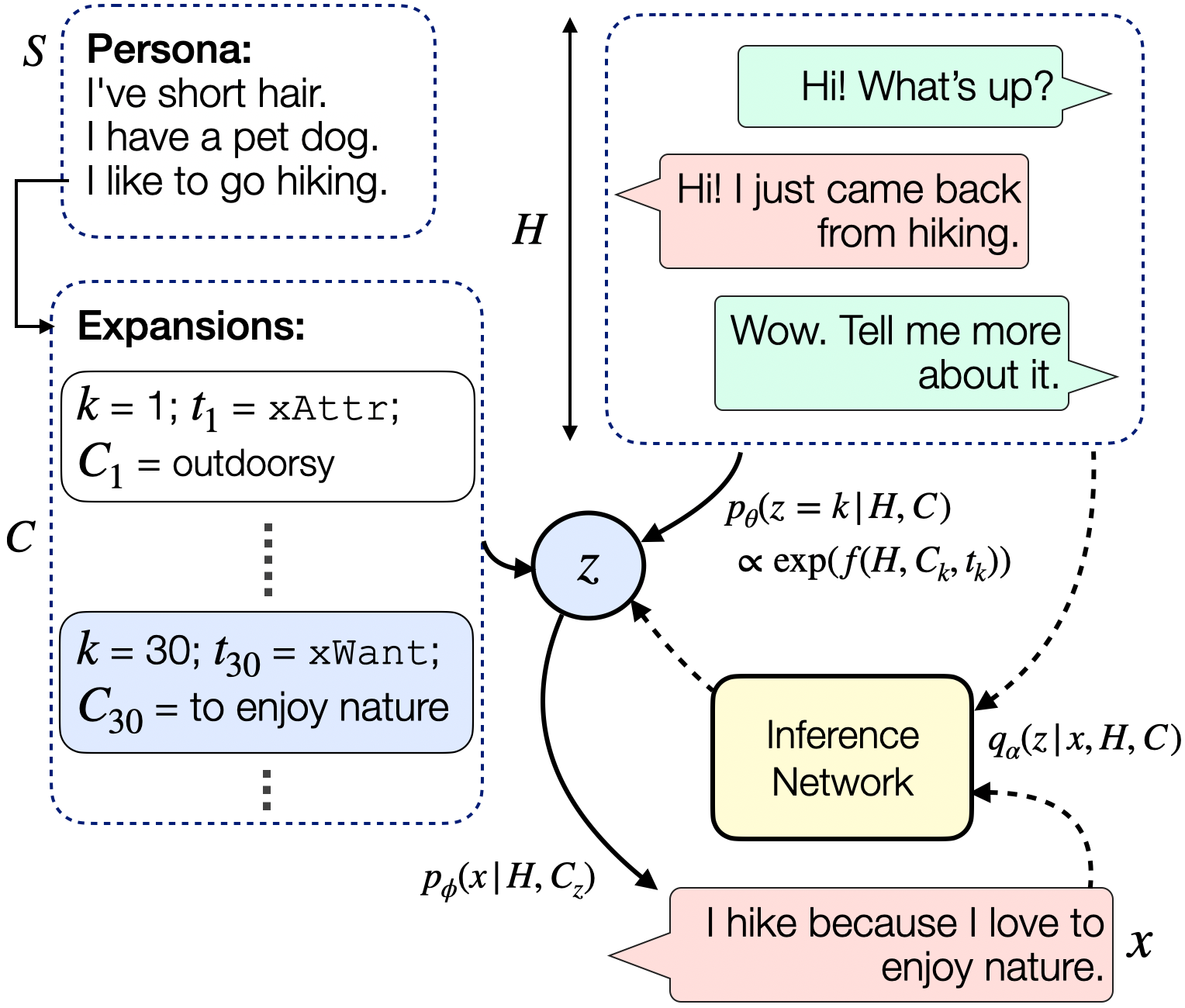}
\centering
\caption{\compac~samples a persona sentence from the prior and generates the response conditioned on the dialog context and sampled persona. The inference network is used only during training.
}
\label{fig:model}
\end{figure}

The dialog history $H$ can hold cues regarding which persona sentence might be applicable given the context. For example, in \Cref{fig:model} the historical context suggests that `following fashion trends' can be a consequence of `being fashionable'.

We encode both the dialog history $H$ and persona sentence $C_k$ by averaging RoBERTa subword embeddings \cite{DBLP:journals/corr/abs-1907-11692} as $e(H)$ and $e(C_k)$. We use an implementation from HuggingFace for RoBERTa\footnote{\url{https://huggingface.co/transformers/model_doc/roberta.html}} with \texttt{roberta-base} as the pretrained model.
Then we parameterize the prior $p_\theta(z|H, C)$ as a log-linear model with the following features:

\newpara{Dialog history}We obtain $f_1(H, C_k)$: a scalar feature using a bilinear product $\langle e(H), e(C_k)\rangle$ to align the persona sentences with the dialog history.

\newpara{Expansion types}Each $k$-th persona expansion corresponds to an expansion type $t_k$. In the case of COMET, these types are the nine commonsense relations provided by ATOMIC (see \Cref{sec:comet}). For paraphrased expansions, we annotate each 
as type \texttt{paraphrase} and the original persona sentences 
as
\texttt{original}.
We consider two additional features with expansion types:
(a) $f_2(t_k)$ that represents a global preference over the relation type embedded via a type embedding layer;
and
(b) $f_3(t_k, H)$ that appends the expansion type embedding with dialog history encoding $e(H)$, followed by a linear layer to obtain a real-valued score for history-specific preference over the 
expansion type.

The dimension of the expansion type embedding was set to 5. Finally, the prior model can be represented concisely as $ p_\theta(z=k|H, C) \propto \exp( f(H, C_k, t_k) )$, where $f(H, C_k, t_k)$ is the sum $\lambda_1*f_1(H,C_k) + \lambda_2*f_2(t_k) + \lambda_3*f_3(t_k, H)$ 
with $\lambda_i$'s are trainable parameters.

\subsection{Generator Network}
Following prior work \cite{DBLP:journals/corr/abs-1901-08149}, we use pre-trained GPT2 \cite{Radford2018ImprovingLU} (Transformer with 12 layers, 768 hidden size, 12 heads--- \texttt{gpt2-small}\footnote{\url{https://github.com/huggingface/transfer-learning-conv-ai}}) to generate dialog responses given the dialog history $H$, with the selected persona sentence $C_z$ prepended to it.
In the case of $C_z$ being the null persona, an empty string is prepended.
We further append the target response $x$ to the combined context ($C_z; H$), and feed the sequence to GPT2, after tokenization.
To distinguish between persona tokens, history tokens, and target response tokens, we use segment indicators---\{\texttt{Persona}, \texttt{Speaker1}, \texttt{Speaker2}\}---for which corresponding embeddings are learned via a separate segment embedding layer in the model.
We add the segment embedding to the corresponding token embedding in the model input layer. 
To obtain the conditional likelihood $p_{\phi}(x | H, C_z)$, we only consider the target tokens for cross-entropy loss calculation.

\citet{DBLP:journals/corr/abs-1901-08149} also leveraged incorrect responses given a dialog history from \personachat~as negative samples in an auxiliary loss to encourage the correct candidate to obtain the highest likelihood compared to the incorrect ones. However, we did not find any improvement using this loss in \compac.


\subsection{Learning and Inference}
Our training data $\mathcal{D}$ consists of instances of dialog history $H$ and ground truth dialog responses $x$. We train our model parameters $\theta$ and $\phi$ to maximize the likelihood of the target dialog response $x$ given the dialog history: $\log p(x|H, C;\theta,\phi)$ totalled over $\mathcal{D}$.
Since the discrete random variable $z$ is unobserved in the training data, we must marginalize over $z$ to compute the desired likelihood $ p(x|H;\theta,\phi)$: 
\[
    \log p(x|H;\theta,\phi) = \log \mathbb{E}_{z \sim p_{\theta}(z|H) } [ p_{\phi}(x|z,H) ];
\]
where we drop $C$ from the conditionals for simplicity. \\

\newpara{Inference Network}
Note that the
number of persona expansions is typically in the 
range 150-250, and thus  it is computationally  expensive to marginalize over the entire selection space of $z$ during training. We instead optimize a variational lower bound (ELBO) of $\log p(x|H;\theta,\phi)$ given as
\begin{align*}
&\mathbb{E}_{z \sim q_{\alpha}(z|H) } [ \log p_{\phi}(x|z,H) ] \\
&- KL(q_\alpha(z|x,H)||p_{\theta}(z|H)),    
\end{align*}
where we use the inference network $q_{\alpha}(z|x, H)$ to compute the approximate posterior \cite{DBLP:journals/corr/KingmaW13}. In our initial experiments, we observe that using an inference network leads to better perplexity values than using samples from the prior.  

The architecture of the inference network is similar to that of the prior network, a log-linear model. Along with the features related to dialog history and expansion types, we additionally include another scalar feature: a bilinear product $\langle x, C_k\rangle$ between the encoded persona and ground truth response $x$ encoded with RoBERTa embeddings to align the persona choice according to the target utterance. \\


\newpara{Optimization}
The parameters of the generator network ($\phi$) and prior network ($\theta$) can be trained directly via back-propagation.
Since $z$ is a discrete latent variable, we use REINFORCE \cite{williams1992simple} to train the inference network parameters $\alpha$.  However, the REINFORCE estimator often suffers from high variance. To reduce the variance, we found it useful to (1) use a moving average baseline \cite{DBLP:conf/nips/ZhaoHNS11}; and (2) regularize the prior network by penalizing the entropy of the output categorical distribution. 
To avoid KL mode collapse, we use KL-annealing \cite{bowman2016generating} where we linearly increase the weight of the KL term beginning from 0 to 1 as training progresses. \\

\newpara{Decoding}
At decoding time, we first sample $k$ from the prior 
$p_\theta(z|H, C)$, 
and then $C_k$ is fed to the generator network.
Following previous work \cite{DBLP:journals/corr/abs-1901-08149}, we use nucleus sampling \cite{DBLP:conf/iclr/HoltzmanBDFC20}
(with $p=0.95$) to decode the final response from the probabilities produced by the generator.
We also found that high-temperature sampling from the prior often leads to more diverse generation.



\section{Experiments}

We conduct our experiments based on the following desiderata: (1) Do persona expansions help to generate high quality
and diverse responses? (2) Does \compac~achieve accurate persona grounding given a dialog context? (3) Does \compac~enable persona-consistent and controllable generation? Hyperparameter details are in Appendix \S A.


\begin{table*}[t!]
\small
\centering
\begin{minipage}{0.56\textwidth}
\centering
\begin{tabular}{@{}lccccc@{}}
\toprule
\bf System                   & \bf PPL  & \bf BLEU-1 & \bf BLEU-2  & \bf D-1 & \bf D-2 \\ \midrule
\multicolumn{6}{@{}l}{\textbf{Original}}\\
Per-CVAE \shortcite{DBLP:conf/ijcai/SongZCWL19}          
& 48.37        & 0.19 & 0.11        & 0.03       & 0.21      \\
LIC + KS \shortcite{DBLP:conf/ijcai/LianXWPW19}          
& 30.50        & 0.18 & 0.07        & 0.07       & 0.24      \\
GPT2 \shortcite{DBLP:journals/corr/abs-1901-08149}         
& 21.46        & 1.42  & 0.78     & 0.05       & 0.11   \\
\compac-original          
& 19.56     & 3.24  & 1.31  & 0.15       & 0.25       \\ \midrule
\multicolumn{6}{@{}l}{\textbf{Paraphrased}}\\
GPT2-revised        & 21.01     & 1.54  & 0.97   & 0.13       & 0.25       \\
GPT2-paraphrase        & 21.57     & 1.61  & 0.86   & 0.16       & 0.35       \\
\compac-revised          
& 18.12     & 3.52  &  0.99  & 0.48       & 0.65       \\
\compac-paraphrase       
& 17.09     & 3.83  & \bf 1.87   & 0.56       & 0.85       \\ \midrule
\multicolumn{6}{@{}l}{\textbf{COMET}}\\
GPT2-COMET        & 21.12     & 1.62  & 0.81   & 0.21       & 0.39       \\
\compac         & \bf 16.21       & \bf 4.12   & 1.82  & \bf 0.87  & \bf 1.07 
\\\bottomrule
\end{tabular}
\caption{Dialog quality metrics on 
the
\personachat~test set. PPL=Perplexity, D-1/2=\% of distinct uni- and bi-grams.}
\label{tab:metrics}
\end{minipage}%
\hfill
\begin{minipage}{0.40\textwidth}
\centering
\begin{tabular}{@{}l@{}}
\toprule
\begin{tabular}{@{}l@{}}
\textbf{Persona:} \\
I enjoy listening to classical music.\\
I'm a Hindu. \\
My favorite color is red.\\ \midrule
\midrule
\textbf{User:} Hi, recently I have got interests in religion.\\ \midrule
\textbf{GPT2} \shortcite{DBLP:journals/corr/abs-1901-08149}\textbf{:}
Hi! How are you? \\ \midrule
\textbf{\compac-original:}
I'm a Hindu. \\ \midrule
\textbf{\compac-revised:}
Hi! I am a Hindu too. \\ \midrule
\textbf{\compac-paraphrase:}
That's great. I am \\religious. \\ \midrule
\textbf{\compac:}
That's great. I go to temple regularly \\and learn about Hinduism.\\ \bottomrule
\end{tabular}
\end{tabular}
\caption{Sample generations by different models. More examples are in Appendix \S C.}
\label{tab:sample}

\end{minipage}
\end{table*}

\begin{table*}[t!]
\small
\centering
\begin{tabular}{lcc|cc|cc|rc|rc|cc}
\toprule
\bf \compac~vs.        & \multicolumn{2}{c|}{\bf GPT2 \shortcite{DBLP:journals/corr/abs-1901-08149}} & \multicolumn{2}{c|}{\bf LIC + KS \shortcite{DBLP:conf/ijcai/LianXWPW19}} & \multicolumn{2}{c|}{\bf GPT2-COMET} & \multicolumn{2}{c|}{\bf \compac-og} & \multicolumn{2}{c|}{\bf \compac-par} & \multicolumn{2}{c}{\bf Gold} \\ \midrule
\bf Metric $\downarrow$      & win          & loss   & win          & loss       &  win            & loss           &  win         &  loss         &  win           &  loss          &  win        &  loss        \\ \midrule
Fluency      &          \bf 81.2*    &   5.1    & \bf 83.2* & 6.7        &    \bf 90.5*            &   2.3              &  \bf 68.0           &    26.0          &     \bf 65.0          &  19.4              &    40.1        &   \bf 42.5          \\ 
Engagement &      \bf 90.5*        &   3.3      & \bf 87.4  & 5.9         &     \bf 97.6*           &    0.5            &    \bf 86.5*         &      10.5        &  \bf 81.5*            &   10.5            &   \bf 62.1*         &      30.5       \\ 
Relevance    &      \bf 78.2*        &   4.8    & \bf 78.0* & 7.7       &        \bf 93.2*        &       1.8         &   \bf 65.5*          &     18.5         &  \bf 62.1             &      15.6         &      32.8      &  \bf 54.6* \\ \bottomrule
\end{tabular}
\caption{Pairwise comparison between responses generated by \compac~ vs.~responses generated by other baselines (og: original, par: paraphrase) as well as 
the
Gold response. All numbers are in percentages with \textbf{bold} indicates the highest. Ties are not shown. Entries with * denote significance with $p < 0.05 $ from bootstrap tests on 1000 subsets of size 50.}
\label{tab:human1}
\end{table*}

\subsection{Baselines}
\label{sec:baseline}

To demonstrate the efficacy of \compac, we compare it with three competitive baselines on 
the
\personachat~dataset:
\begin{enumerate}
    \item \textbf{Per-CVAE:} A CVAE model that exploits persona sentences for diverse generation with an external memory \cite{DBLP:conf/ijcai/SongZCWL19}
    \item \textbf{LIC + KS:} The best performing transformer model (Lost in Conversation i.e.,~LIC) in terms of human evaluation in the ConvAI2 NeurIPS competition \cite{DBLP:journals/corr/abs-1902-00098} combined with a knowledge-selection (KS) mechanism \citet{DBLP:conf/ijcai/LianXWPW19}  to achieve 
    state-of-the-art results on \personachat;
    \item \textbf{GPT2:} Finetuned GPT2 on \personachat~just by concatenating all persona sentences along with dialog history \citep{DBLP:journals/corr/abs-1901-08149}  to obtain the best automatic metric in the ConvAI2 competition.
\end{enumerate}

A minimal version of \compac~is also considered, \textbf{\compac-original}, which only uses the original persona, for a direct comparison with other model architectures that only use the original persona.
Furthermore, to justify the choice of fine-grained persona grounding for an effective utilization of persona expansions, we also consider baseline versions of GPT2 trained with each of the expansion strategies: \textbf{GPT2-revised}, \textbf{GPT2-paraphase}, and \textbf{GPT2-COMET}. To show that \compac~can work with persona expansions derived from various sources, we compare with versions of \compac~trained with paraphrase-based expansions: \textbf{\compac-revised} and \textbf{\compac-paraphrase}. By default, \compac~indicates it is trained with COMET expansions.

\subsection{Comparison of Dialog Quality}
\label{sec:d_quality}
We measure perplexity for language modeling performance, and
BLEU-1 \cite{DBLP:conf/acl/PapineniRWZ02} and BLEU-2 \cite{DBLP:conf/cvpr/VedantamZP15} scores between generated and gold utterances to measure the fidelity of the generated responses.
Given our goal of generating
engaging responses
with novel information, we deem it important to consider the diversity in the generated responses
which we measure using D-1 and D-2 (percentage of distinct uni- and bi-grams respectively) \cite{DBLP:conf/naacl/LiGBGD16}.

\Cref{tab:metrics} shows 
that \compac~outperforms three competitive baselines when trained on the original persona in all quality metrics indicating the efficacy of our architecture.
Moreover, when combined with persona expansions, we observe a modest 3-8 point decrease in perplexity and a large improvement in both BLEU and diversity scores which 
confirms
that \compac~successfully leverages the persona expansions to improve dialog quality.
\compac~trained with COMET expansions achieves the best performance both in terms of fidelity and diversity which shows that COMET expansions help the model to respond to implicit context with commonsense 
and
to explore novel information.
But with revised personas, we find that both \compac~and GPT2 provide marginal performance gains, mirroring the observation from \cite{DBLP:conf/acl/KielaWZDUS18}.
Finally we observe gradual degradation in performance when we trivially finetune GPT2 with paraphrase and COMET expansions. Note that GPT-2 could have implicitly learned to focus on a single persona attribute. However, the proposed \compac~model performs better  suggesting that fine-grained persona grounding acts as a useful inductive bias in effectively utilizing larger expansion sets.

\subsection{Human Evaluation for Dialog Generation} 
\label{sec:human_eval}
Automatic evaluation of dialog systems
is
still
notoriously 
unreliable
\cite{DBLP:conf/emnlp/LiuLSNCP16, DBLP:conf/emnlp/NovikovaDCR17} and
such systems should be
evaluated by human users. 
Hence, we perform pairwise comparisons between responses generated our best system, \compac~trained on COMET expansions, and responses generated by four strong baselines: GPT2, GPT2-COMET, \compac-original, 
\compac-paraphrase (the best \compac~model with paraphrase expansions).
We also consider the gold responses for comparison.
We conduct a human evaluation with 100 test examples
on three aspects 
critical for
practical use:
(1) \textbf{Fluency} measures whether the generated output is fluent (in English);
(2) \textbf{Engagement} measures whether the generated response is engaging or interesting;
and 
(3) \textbf{Relevance} measures whether the generated output is relevant with respect to the dialog history.
More details of the evaluation
are in Appendix \S B.

\Cref{tab:human1} shows that human annotators found responses generated by \compac~trained with COMET expansions 
more engaging as compared to responses from all the baselines as well as the gold responses by statistically significant margins. This confirms our hypothesis that COMET expansions were helpful in adding novel content. 
Human judges also found that despite a significant drop in perplexity, \compac~was not more fluent than \compac-original and \compac-paraphrase with statistical significance, indicating similar language modeling performance.
We find the inter-annotator agreement, as measured by 
Cohen's
kappa \cite{cohen1960coefficient}, for fluency, engagement, and relevance were 0.62, 0.71, and 0.73 respectively.

\begin{table}[t!]
\small
\centering
\begin{tabular}{@{}lccc@{}}
\toprule
\bf \multirow{2}{*}{\bf System}       & \multicolumn{2}{c}{\bf Persona Entailment} &  \bf Human  \\ 
& \bf \hspace{-2mm}Prior & \hspace{-2mm} \bf Inference Network & \bf eval. \\
\midrule
\multicolumn{4}{@{}l}{\textbf{Original}}\\
\compac-original\hspace{-2mm}  
& 25.5 & \hspace{-2mm}79.3 & -- \\ \midrule
\multicolumn{4}{@{}l}{\textbf{Paraphrased}}\\
\compac-revised\hspace{-2mm} & 20.6 & \hspace{-2mm}78.9   &  40.6 \\ 
\compac-paraphrase\hspace{-2mm}  & 27.8 & \hspace{-2mm}87.3   & 67.8  \\ \midrule
\multicolumn{4}{@{}l}{\textbf{COMET}}\\
\compac\hspace{-2mm} & \bf 37.9 & \hspace{-2mm}\bf 96.4    &   \bf 87.3\\ \bottomrule
\end{tabular}
\caption{Assessment of persona grounding with and without inference network
using the DNLI entailment set. Human evaluation (eval.)~was conducted to measure the relevance when an expanded persona is chosen--all entries are statistically significant.
}
\label{tab:interpret}
\end{table}

\subsection{Fine-grained Persona Grounding}
\label{sec:interpret}

Next
we want to investigate the extent of \compac's ability to ground the response generation with a fine-grained persona choice as a probing experiment.
Specifically, we want to measure whether our model can choose a coherent persona from the available persona sentences given the dialog context. Note that in persona-grounded chitchat, not all utterances are tied to a personas and could be purely based on dialog context. We find that 44\% of the time the model 
selects the
null persona ($\varnothing$) and conditions only on the dialog history. To assess the persona grounding for the remaining (56\%) utterances, we perform (a) a persona entailment experiment, and (b) human evaluation.

\newpara{Persona Entailment}
We adapt the Dialogue Natural Language Inference (DNLI) dataset \cite{DBLP:conf/acl/WelleckWSC19} and collect persona-utterance pairs that belong to an \textit{entailment} relation.
This results in a subset of 4,613 utterances with associated ground truth persona sentences in our test set.
Next, we obtain a persona sentence by performing $\operatorname{argmax}$ over the prior $p_{\theta}(z|H, C)$ as well as the inference network $q_\alpha(z|x, H, C)$ from our \compac~models and calculate accuracy with the ground truth persona.
For models that use 
expanded personas,
we track the original persona from the retrieved expansion for accuracy calculation.
\Cref{tab:interpret} shows that \compac~with COMET achieves the most accurate persona grounding suggesting that inference networks can approximate the true posterior better when a commonsense persona is available for grounding.
In the case of the prior, a better entailment accuracy than random chance ($1/5$) confirms our choice of the history-conditioned prior network rather than a uniform prior.

\newpara{Human Evaluation}
Since DNLI does not entail expanded personas, we conduct a human evaluation to judge the relevance of a chosen persona \textit{expansion} sampled from the inference network.
Specifically, we ask: 
\textit{Is this knowledge relevant to the given dialog history?}---with options as `Yes', `No', and `Uncertain'---and with 100 examples (more in Appendix \S B) for each \compac~variant that uses expanded personas. The inter-annotator agreement, as measured by Cohen's kappa was 0.76.
Again, \Cref{tab:interpret} shows that models with COMET expansions can choose the most relevant persona sentence which corroborates our claim in persona entailment experiments.
On average, we noticed that \compac~with COMET expansions prefers to choose expanded personas 87\% of the time out of all non-null persona choices.
This reduces to 62\% in the case \compac-paraphrase. In contrast, \compac-revised tends to select an original persona over an expansion more often.

\begin{table}[t!]
\small
\centering
\begin{tabular}{@{}lcccc@{}}
\toprule
\bf \multirow{2}{*}{\bf System} & \multicolumn{3}{c}{\bf Unigram Overlap} & \bf BERT \\
& \bf Recall & \bf Precision & \bf F1 & \bf Score\\ 
\midrule
\multicolumn{5}{@{}l}{\textbf{Original}}\\
LIC + KS \shortcite{DBLP:conf/ijcai/LianXWPW19}\hspace{-4mm} 
& 10.4 & 34.2  & 15.3   & --     \\
\compac-original\hspace{-4mm}    
& 14.9 & 39.1 & 21.6   & 57.2\\ \midrule
\multicolumn{5}{@{}l}{\textbf{Paraphrased}}\\
\compac-revised\hspace{-4mm}      & 15.2 & 40.3   & 22.1   & 58.1 \\ 
\compac-paraphrase\hspace{-4mm}      & 17.8 & 42.2 & 25.1 & 72.9\\ \midrule
\multicolumn{5}{@{}l}{\textbf{COMET}}\\
\compac\hspace{-4mm}    & \bf 21.4 & \bf 48.9 &  \bf 29.8  & \bf 78.8 \\ \bottomrule
\end{tabular}
\caption{Conditional generation performance on the \personachat~test set to show the similarity between generated responses and grounding persona sentences. We omit GPT2-based models since they do not select a particular persona sentence for grounding.
}
\label{tab:overlap}
\end{table}


\subsection{Controllable Generation}
Controllable generation of persona-grounded dialog can help to generalize the dialog agent to newer persona details just by changing the grounding in the conditional generator.
While controllable text generation with a desired attribute has gained interest recently \cite{DBLP:conf/iclr/DathathriMLHFMY20, DBLP:journals/corr/abs-1901-07129}, we investigate the possibility of controlling generation with a desired persona and
measure the performance of the conditional generator.
For this, we observe
a set of knowledge overlap metrics---the unigram recall/precision/F1 scores--from \citet{DBLP:conf/iclr/DinanRSFAW19} and BERT score \cite{DBLP:conf/iclr/ZhangKWWA20} for semantic similarity between the generated responses and the persona retrieved.
\Cref{tab:overlap} shows that conditional generation is strongest when \compac~is trained with COMET suggesting commonsense expansions are more appropriate to the dialog context in influencing the response generation. 

Next, we create a diagnostic dataset of 100 examples where we manually edit the persona by changing an entity in a persona sentence or swapping the selected persona expansion with another relevant one (See examples in \Cref{tab:control}) to directly measure controllability in response generation.
We observe that \compac~can successfully reflect the entity-change in the generated response based on the change in the persona grounding 86\% of the time.
For a swapped persona expansion, we also see a higher BERT score (74.6) between the edited persona and newly generated response as opposed to a lower score (46.2) with the unedited persona.
Together
with the qualitative examples in \Cref{tab:control} this 
suggests that \compac~supports controllable generation with contextually modified personas.

\begin{table}[t!]
\small
\centering
\begin{tabular}{@{}ll@{}}
\toprule
\bf Performance & \hspace{-2mm}\bf Example \\ \midrule
\begin{tabular}{@{}l@{}}
Presence of\\changed entity\\
\textbf{86\%}
\end{tabular} &
\hspace{-2mm}\begin{tabular}{@{}l@{}}
\textbf{Changing the key entity} \\
Before: My favorite color is red \\
After: My favorite color is \textbf{green} \\ \midrule
Conversation:\\
User: What is your favorite color?\\
Bot:  My favorite color is \textbf{green} \hspace{4em}              
\end{tabular}
\\ \midrule \midrule
\begin{tabular}{@{}l@{}}
BERT score with\\unedited persona:\\
46.2
\\edited persona:\\
\textbf{74.6}
\end{tabular} &
\hspace{-2mm}\begin{tabular}{@{}l@{}}
\textbf{Swapping with another expansion} \\
Before: I want to swim in the ocean \\
After: I want to buy a beach \textbf{umbrella} \\ \midrule
Conversation:\\
User: What do you do at beaches?\\
Bot:  I will buy an \textbf{umbrella} at the beach
\end{tabular}
\\ \bottomrule
\end{tabular}
\caption{Controlled generation with edited persona}
\label{tab:control}
\end{table}

\subsection{Qualitative Analysis}

\begin{table*}
\small
\centering

\begin{tabular}{@{}cllll@{}}
\toprule
\bf Verdict & \bf Persona & \bf Dialog History & \bf Persona Choice ($z$) & \bf Response \\ \midrule
\begin{tabular}{@{}l@{}}
\begin{turn}{90} 
Uses
\end{turn}
\begin{turn}{90} 
Original
\end{turn}
\begin{turn}{90} 
Persona
\end{turn}
\end{tabular} &
\begin{tabular}{@{}l@{}}
1. I'm the youngest of five.\\
2. I work at the hospital as a nurse.\\
3. I've pink hair. 
\end{tabular} &
\begin{tabular}{@{}l@{}}
What do you do\\for work?
\end{tabular}
& 
\begin{tabular}{@{}l@{}}
I work at the \\hospital as a nurse.
\end{tabular}
& 
\begin{tabular}{@{}l@{}}
I am a nurse.
\end{tabular}
\\ \midrule \midrule
\begin{tabular}{@{}l@{}}
\begin{turn}{90} 
Uses
\end{turn}
\begin{turn}{90} 
Expanded
\end{turn}
\begin{turn}{90} 
Persona
\end{turn}
\end{tabular} &
\begin{tabular}{@{}l@{}}
1. I just want to have fun with my friends.\\
2. I don't drink or do drugs or anything.\\
3. I am 19 and I cannot wait \\to move out of my parents home.
\end{tabular}& 
Are you enjoying life?
& 
\begin{tabular}{@{}l@{}}
I just want to have fun \\with my friends $\rightarrow$ \\wants to have a party
\end{tabular}& 
\begin{tabular}{@{}l@{}}
Not really.\\I want to have a\\party.
\end{tabular}
\\ \midrule \midrule
\begin{tabular}{@{}l@{}}
\begin{turn}{90} 
Uses
\end{turn}
\begin{turn}{90} 
Improper
\end{turn}
\begin{turn}{90} 
Persona
\end{turn}
\end{tabular}
 &
\begin{tabular}{@{}l@{}}
1. I make a million dollars a year.\\
2. I'm married and have three kids.\\
3. I'm a baseball player.\\
\end{tabular} &
\begin{tabular}{@{}l@{}}
I find it hard to support\\my family working at\\a bar. What about you? 
\end{tabular} & 
Null persona ($\varnothing$) & I enjoy my life.\\

\bottomrule
\end{tabular}
\caption{Examples showing correct and incorrect persona choices in various dialog contexts by \compac~model. It shows that \compac~is capable of choosing a correct persona sentence (original or expanded) but sometimes the prior network fails to sample an appropriate one (third case).}
\label{tab:error}
\end{table*}

\Cref{tab:sample} shows responses from different models
for a sample dialog context.
Qualitatively, we find that \compac~with COMET expansions
responds to the context with commonsense using novel content from a commonsense expansion (being Hindu $\rightarrow$ to learn about Hinduism), where other responses remain generic or incoherent.
In \Cref{tab:error}, we illustrate responses generated by the \compac~model along with the underlying persona choice sampled from the prior network. Cases show that \compac~successfully chooses an original or an expanded persona sentence, as appropriate, but also defaults to the null persona ($\varnothing$) that leads to a bland response.

\section{Related Works}
Building personalized dialog agents has been a popular task recently, thanks to \citet{DBLP:conf/acl/KielaWZDUS18} who extensively studied the task with a new dataset
\personachat, later as a form of a challenge \cite{DBLP:journals/corr/abs-1902-00098}, where the dialog agent is seeded with a predefined persona in 
the
form 
of
multiple sentences of textual description,
mirroring a casual human conversation which many times draws snippets from individual personal experiences and facts.
Recent works focus on improving persona-grounded dialog generation performance \cite{DBLP:journals/corr/abs-1901-08149, DBLP:conf/emnlp/MazareHRB18, DBLP:conf/acl/BaoHWLW19} as well as persona consistency in generated dialog \cite{DBLP:conf/acl/WelleckWSC19, DBLP:journals/corr/abs-1911-03860, DBLP:journals/corr/abs-1911-05889}.
\citet{DBLP:conf/acl/BaoHWLW19} proposed a reinforcement-learning-based framework that promoting informativeness and persona-consistency via personal knowledge exchange.
\citet{DBLP:journals/corr/abs-2002-02153} focused on using plausible topical keywords related to the available persona facts using a neural topic model to explore beyond the given knowledge, possibly closest to our work.
We rather focus on obtaining commonsense implications of the given persona in the form of text snippets that are more expressive than topical keywords.

Persona-grounded dialog generation is a special case of knowledge-grounded dialog generation.
Knowledge grounding in dialog has many real-world applications that are well-studied in recent literature \cite{DBLP:conf/ijcai/ZhouYHZXZ18, DBLP:conf/aaai/GhazvininejadBC18, DBLP:conf/iclr/DinanRSFAW19, DBLP:journals/corr/abs-1910-13461}.
In this work we use fine-grained grounding/selection on persona which performed better than encoding the entire persona for each response. Such fine-grained selection has been found useful in prior works on text generation such as dialog \cite{DBLP:conf/ijcai/LianXWPW19} and image captioning \cite{DBLP:conf/emnlp/JhamtaniB18}.
For dialog generation, a contextual knowledge selection has been successfully applied in prior works \cite{DBLP:conf/emnlp/ParthasarathiP18}.
Specifically, \citet{DBLP:conf/acl/ZhaoZE17} and later \citet{DBLP:conf/ijcai/SongZCWL19} proposed a conditional-VAE framework to learn latent context given the dialog history to guide knowledge selection.

Finally, few recent works focused on augmenting grounding with commonsense knowledge with successful applications in open-domain topical dialog generation \cite{DBLP:conf/aaai/GhazvininejadBC18, DBLP:conf/acl/MoonSKS19}, story generation \cite{DBLP:conf/emnlp/MaoMMC19} and sarcasm generation \cite{DBLP:journals/corr/abs-2004-13248}. In this work, we extend this effort into persona-grounded dialog generation via augmenting grounding persona with commonsense knowledge.
 
\section{Conclusion}
In this work, we showed that expanding persona sentences with commonsense helps a dialog model to generate high-quality and diverse persona-grounded responses.
Moreover, we found that \textit{fine-grained} persona grounding is crucial to effectively condition on a large pool of commonsense persona expansions, which further provided additional controllability in conditional generation.

While our expansions are limited by the performance of COMET or paraphrase systems, we envision future work to train the dialog model end-to-end along with the expansion generation.
As future work, we would like extend the prior network to sample more than one persona sentences by expanding the sample space of the discrete random variable to generate more interesting responses.

\paragraph{Acknowledgements}
We thank Arthur Szlam, Y-Lan Boureau, Michel Galley, Sujoy Paul, and anonymous reviewers for providing valuable feedback on this work. BPM is partly supported by NSF Award \#1750063. HJ is supported in part by a Adobe Research Fellowship. Findings and observations are of the authors only, and do not necessarily reflect the views of the funding agencies. 


\bibliography{emnlp2020}

\begin{thebibliography}{43}
\expandafter\ifx\csname natexlab\endcsname\relax\def\natexlab#1{#1}\fi

\bibitem[{Bao et~al.(2019)Bao, He, Wang, Lian, and
  Wu}]{DBLP:conf/acl/BaoHWLW19}
Siqi Bao, Huang He, Fan Wang, Rongzhong Lian, and Hua Wu. 2019.
\newblock \href {https://doi.org/10.18653/v1/p19-1535} {Know more about each
  other: Evolving dialogue strategy via compound assessment}.
\newblock In \emph{ACL}.

\bibitem[{Bosselut et~al.(2019)Bosselut, Rashkin, Sap, Malaviya,
  {\c{C}}elikyilmaz, and Choi}]{DBLP:conf/acl/BosselutRSMCC19}
Antoine Bosselut, Hannah Rashkin, Maarten Sap, Chaitanya Malaviya, Asli
  {\c{C}}elikyilmaz, and Yejin Choi. 2019.
\newblock \href {https://doi.org/10.18653/v1/p19-1470} {{COMET:} commonsense
  transformers for automatic knowledge graph construction}.
\newblock In \emph{ACL}.

\bibitem[{Bowman et~al.(2016)Bowman, Vilnis, Vinyals, Dai, Jozefowicz, and
  Bengio}]{bowman2016generating}
Samuel Bowman, Luke Vilnis, Oriol Vinyals, Andrew Dai, Rafal Jozefowicz, and
  Samy Bengio. 2016.
\newblock Generating sentences from a continuous space.
\newblock In \emph{SIGNLL}.

\bibitem[{Chakrabarty et~al.(2020)Chakrabarty, Ghosh, Muresan, and
  Peng}]{DBLP:journals/corr/abs-2004-13248}
Tuhin Chakrabarty, Debanjan Ghosh, Smaranda Muresan, and Nanyun Peng. 2020.
\newblock \href {http://arxiv.org/abs/2004.13248} {$ {R}^3$: Reverse, retrieve,
  and rank for sarcasm generation with commonsense knowledge}.
\newblock \emph{CoRR}, abs/2004.13248.

\bibitem[{Cohen(1960)}]{cohen1960coefficient}
Jacob Cohen. 1960.
\newblock A coefficient of agreement for nominal scales.
\newblock \emph{Educational and psychological measurement}, 20(1):37--46.

\bibitem[{Dathathri et~al.(2020)Dathathri, Madotto, Lan, Hung, Frank, Molino,
  Yosinski, and Liu}]{DBLP:conf/iclr/DathathriMLHFMY20}
Sumanth Dathathri, Andrea Madotto, Janice Lan, Jane Hung, Eric Frank, Piero
  Molino, Jason Yosinski, and Rosanne Liu. 2020.
\newblock \href {https://openreview.net/forum?id=H1edEyBKDS} {Plug and play
  language models: {A} simple approach to controlled text generation}.
\newblock In \emph{ICLR}.

\bibitem[{Dinan et~al.(2019{\natexlab{a}})Dinan, Logacheva, Malykh, Miller,
  Shuster, Urbanek, Kiela, Szlam, Serban, Lowe, Prabhumoye, Black, Rudnicky,
  Williams, Pineau, Burtsev, and Weston}]{DBLP:journals/corr/abs-1902-00098}
Emily Dinan, Varvara Logacheva, Valentin Malykh, Alexander~H. Miller, Kurt
  Shuster, Jack Urbanek, Douwe Kiela, Arthur Szlam, Iulian Serban, Ryan Lowe,
  Shrimai Prabhumoye, Alan~W. Black, Alexander~I. Rudnicky, Jason Williams,
  Joelle Pineau, Mikhail Burtsev, and Jason Weston. 2019{\natexlab{a}}.
\newblock \href {http://arxiv.org/abs/1902.00098} {The second conversational
  intelligence challenge (convai2)}.
\newblock \emph{CoRR}, abs/1902.00098.

\bibitem[{Dinan et~al.(2019{\natexlab{b}})Dinan, Roller, Shuster, Fan, Auli,
  and Weston}]{DBLP:conf/iclr/DinanRSFAW19}
Emily Dinan, Stephen Roller, Kurt Shuster, Angela Fan, Michael Auli, and Jason
  Weston. 2019{\natexlab{b}}.
\newblock \href {https://openreview.net/forum?id=r1l73iRqKm} {Wizard of
  wikipedia: Knowledge-powered conversational agents}.
\newblock In \emph{ICLR}.

\bibitem[{Federmann et~al.(2019)Federmann, Elachqar, and
  Quirk}]{DBLP:conf/aclnut/FedermannEQ19}
Christian Federmann, Oussama Elachqar, and Chris Quirk. 2019.
\newblock \href {https://doi.org/10.18653/v1/D19-5503} {Multilingual whispers:
  Generating paraphrases with translation}.
\newblock In \emph{W-NUT@EMNLP}.

\bibitem[{Ghazvininejad et~al.(2018)Ghazvininejad, Brockett, Chang, Dolan, Gao,
  Yih, and Galley}]{DBLP:conf/aaai/GhazvininejadBC18}
Marjan Ghazvininejad, Chris Brockett, Ming{-}Wei Chang, Bill Dolan, Jianfeng
  Gao, Wen{-}tau Yih, and Michel Galley. 2018.
\newblock \href
  {https://www.aaai.org/ocs/index.php/AAAI/AAAI18/paper/view/16710} {A
  knowledge-grounded neural conversation model}.
\newblock In \emph{AAAI}.

\bibitem[{Holtzman et~al.(2020)Holtzman, Buys, Du, Forbes, and
  Choi}]{DBLP:conf/iclr/HoltzmanBDFC20}
Ari Holtzman, Jan Buys, Li~Du, Maxwell Forbes, and Yejin Choi. 2020.
\newblock \href {https://openreview.net/forum?id=rygGQyrFvH} {The curious case
  of neural text degeneration}.
\newblock In \emph{ICLR}.

\bibitem[{Jhamtani and Berg{-}Kirkpatrick(2018)}]{DBLP:conf/emnlp/JhamtaniB18}
Harsh Jhamtani and Taylor Berg{-}Kirkpatrick. 2018.
\newblock \href {https://doi.org/10.18653/v1/d18-1436} {Learning to describe
  differences between pairs of similar images}.
\newblock In \emph{EMNLP}.

\bibitem[{Kingma and Welling(2014)}]{DBLP:journals/corr/KingmaW13}
Diederik~P. Kingma and Max Welling. 2014.
\newblock \href {http://arxiv.org/abs/1312.6114} {Auto-encoding variational
  bayes}.
\newblock In \emph{ICLR}.

\bibitem[{Kong et~al.(2019)Kong, Li, Neubig, Hovy, and
  Yang}]{DBLP:journals/corr/abs-1901-07129}
Xiang Kong, Bohan Li, Graham Neubig, Eduard~H. Hovy, and Yiming Yang. 2019.
\newblock \href {http://arxiv.org/abs/1901.07129} {An adversarial approach to
  high-quality, sentiment-controlled neural dialogue generation}.
\newblock \emph{CoRR}, abs/1901.07129.

\bibitem[{Lewis et~al.(2019)Lewis, Liu, Goyal, Ghazvininejad, Mohamed, Levy,
  Stoyanov, and Zettlemoyer}]{DBLP:journals/corr/abs-1910-13461}
Mike Lewis, Yinhan Liu, Naman Goyal, Marjan Ghazvininejad, Abdelrahman Mohamed,
  Omer Levy, Veselin Stoyanov, and Luke Zettlemoyer. 2019.
\newblock \href {http://arxiv.org/abs/1910.13461} {{BART:} denoising
  sequence-to-sequence pre-training for natural language generation,
  translation, and comprehension}.
\newblock \emph{CoRR}, abs/1910.13461.

\bibitem[{Li et~al.(2016)Li, Galley, Brockett, Gao, and
  Dolan}]{DBLP:conf/naacl/LiGBGD16}
Jiwei Li, Michel Galley, Chris Brockett, Jianfeng Gao, and Bill Dolan. 2016.
\newblock \href {https://doi.org/10.18653/v1/n16-1014} {A diversity-promoting
  objective function for neural conversation models}.
\newblock In \emph{NAACL HLT}.

\bibitem[{Li et~al.(2019)Li, Roller, Kulikov, Welleck, Boureau, Cho, and
  Weston}]{DBLP:journals/corr/abs-1911-03860}
Margaret Li, Stephen Roller, Ilia Kulikov, Sean Welleck, Y{-}Lan Boureau,
  Kyunghyun Cho, and Jason Weston. 2019.
\newblock \href {http://arxiv.org/abs/1911.03860} {Don't say that! making
  inconsistent dialogue unlikely with unlikelihood training}.
\newblock \emph{CoRR}, abs/1911.03860.

\bibitem[{Lian et~al.(2019)Lian, Xie, Wang, Peng, and
  Wu}]{DBLP:conf/ijcai/LianXWPW19}
Rongzhong Lian, Min Xie, Fan Wang, Jinhua Peng, and Hua Wu. 2019.
\newblock \href {https://doi.org/10.24963/ijcai.2019/706} {Learning to select
  knowledge for response generation in dialog systems}.
\newblock In \emph{IJCAI}.

\bibitem[{Liu et~al.(2016)Liu, Lowe, Serban, Noseworthy, Charlin, and
  Pineau}]{DBLP:conf/emnlp/LiuLSNCP16}
Chia{-}Wei Liu, Ryan Lowe, Iulian Serban, Michael Noseworthy, Laurent Charlin,
  and Joelle Pineau. 2016.
\newblock \href {https://doi.org/10.18653/v1/d16-1230} {How {NOT} to evaluate
  your dialogue system: An empirical study of unsupervised evaluation metrics
  for dialogue response generation}.
\newblock In \emph{EMNLP}.

\bibitem[{Liu et~al.(2019)Liu, Ott, Goyal, Du, Joshi, Chen, Levy, Lewis,
  Zettlemoyer, and Stoyanov}]{DBLP:journals/corr/abs-1907-11692}
Yinhan Liu, Myle Ott, Naman Goyal, Jingfei Du, Mandar Joshi, Danqi Chen, Omer
  Levy, Mike Lewis, Luke Zettlemoyer, and Veselin Stoyanov. 2019.
\newblock \href {http://arxiv.org/abs/1907.11692} {Roberta: {A} robustly
  optimized {BERT} pretraining approach}.
\newblock \emph{CoRR}, abs/1907.11692.

\bibitem[{Loshchilov and Hutter(2017)}]{DBLP:journals/corr/abs-1711-05101}
Ilya Loshchilov and Frank Hutter. 2017.
\newblock \href {http://arxiv.org/abs/1711.05101} {Fixing weight decay
  regularization in adam}.
\newblock \emph{CoRR}, abs/1711.05101.

\bibitem[{Mao et~al.(2019)Mao, Majumder, McAuley, and
  Cottrell}]{DBLP:conf/emnlp/MaoMMC19}
Huanru~Henry Mao, Bodhisattwa~Prasad Majumder, Julian~J. McAuley, and
  Garrison~W. Cottrell. 2019.
\newblock \href {https://doi.org/10.18653/v1/D19-1615} {Improving neural story
  generation by targeted common sense grounding}.
\newblock In \emph{EMNLP}.

\bibitem[{Mazar{\'{e}} et~al.(2018)Mazar{\'{e}}, Humeau, Raison, and
  Bordes}]{DBLP:conf/emnlp/MazareHRB18}
Pierre{-}Emmanuel Mazar{\'{e}}, Samuel Humeau, Martin Raison, and Antoine
  Bordes. 2018.
\newblock \href {https://doi.org/10.18653/v1/d18-1298} {Training millions of
  personalized dialogue agents}.
\newblock In \emph{EMNLP}.

\bibitem[{Moon et~al.(2019)Moon, Shah, Kumar, and
  Subba}]{DBLP:conf/acl/MoonSKS19}
Seungwhan Moon, Pararth Shah, Anuj Kumar, and Rajen Subba. 2019.
\newblock \href {https://doi.org/10.18653/v1/p19-1081} {Opendialkg: Explainable
  conversational reasoning with attention-based walks over knowledge graphs}.
\newblock In \emph{ACL}.

\bibitem[{Novikova et~al.(2017)Novikova, Dusek, Curry, and
  Rieser}]{DBLP:conf/emnlp/NovikovaDCR17}
Jekaterina Novikova, Ondrej Dusek, Amanda~Cercas Curry, and Verena Rieser.
  2017.
\newblock \href {https://doi.org/10.18653/v1/d17-1238} {Why we need new
  evaluation metrics for {NLG}}.
\newblock In \emph{EMNLP}.

\bibitem[{Papineni et~al.(2002)Papineni, Roukos, Ward, and
  Zhu}]{DBLP:conf/acl/PapineniRWZ02}
Kishore Papineni, Salim Roukos, Todd Ward, and Wei{-}Jing Zhu. 2002.
\newblock \href {https://www.aclweb.org/anthology/P02-1040/} {Bleu: a method
  for automatic evaluation of machine translation}.
\newblock In \emph{ACL}.

\bibitem[{Parthasarathi and Pineau(2018)}]{DBLP:conf/emnlp/ParthasarathiP18}
Prasanna Parthasarathi and Joelle Pineau. 2018.
\newblock \href {https://doi.org/10.18653/v1/d18-1073} {Extending neural
  generative conversational model using external knowledge sources}.
\newblock In \emph{EMNLP}.

\bibitem[{Radford(2018)}]{Radford2018ImprovingLU}
Alec Radford. 2018.
\newblock Improving language understanding by generative pre-training.

\bibitem[{Sap et~al.(2019)Sap, Bras, Allaway, Bhagavatula, Lourie, Rashkin,
  Roof, Smith, and Choi}]{DBLP:conf/aaai/SapBABLRRSC19}
Maarten Sap, Ronan~Le Bras, Emily Allaway, Chandra Bhagavatula, Nicholas
  Lourie, Hannah Rashkin, Brendan Roof, Noah~A. Smith, and Yejin Choi. 2019.
\newblock \href {https://doi.org/10.1609/aaai.v33i01.33013027} {{ATOMIC:} an
  atlas of machine commonsense for if-then reasoning}.
\newblock In \emph{AAAI}.

\bibitem[{Song et~al.(2019{\natexlab{a}})Song, Zhang, Hu, and
  Liu}]{DBLP:journals/corr/abs-1911-05889}
Haoyu Song, Wei{-}Nan Zhang, Jingwen Hu, and Ting Liu. 2019{\natexlab{a}}.
\newblock \href {http://arxiv.org/abs/1911.05889} {Generating persona
  consistent dialogues by exploiting natural language inference}.
\newblock \emph{CoRR}, abs/1911.05889.

\bibitem[{Song et~al.(2019{\natexlab{b}})Song, Zhang, Cui, Wang, and
  Liu}]{DBLP:conf/ijcai/SongZCWL19}
Haoyu Song, Weinan Zhang, Yiming Cui, Dong Wang, and Ting Liu.
  2019{\natexlab{b}}.
\newblock \href {https://doi.org/10.24963/ijcai.2019/721} {Exploiting persona
  information for diverse generation of conversational responses}.
\newblock In \emph{IJCAI}.

\bibitem[{Vedantam et~al.(2015)Vedantam, Zitnick, and
  Parikh}]{DBLP:conf/cvpr/VedantamZP15}
Ramakrishna Vedantam, C.~Lawrence Zitnick, and Devi Parikh. 2015.
\newblock \href {https://doi.org/10.1109/CVPR.2015.7299087} {Cider:
  Consensus-based image description evaluation}.
\newblock In \emph{CVPR}.

\bibitem[{Welleck et~al.(2019)Welleck, Weston, Szlam, and
  Cho}]{DBLP:conf/acl/WelleckWSC19}
Sean Welleck, Jason Weston, Arthur Szlam, and Kyunghyun Cho. 2019.
\newblock \href {https://doi.org/10.18653/v1/p19-1363} {Dialogue natural
  language inference}.
\newblock In \emph{ACL}.

\bibitem[{Williams(1992)}]{williams1992simple}
Ronald~J Williams. 1992.
\newblock Simple statistical gradient-following algorithms for connectionist
  reinforcement learning.
\newblock \emph{Machine learning}, 8(3-4).

\bibitem[{Wolf et~al.(2019)Wolf, Sanh, Chaumond, and
  Delangue}]{DBLP:journals/corr/abs-1901-08149}
Thomas Wolf, Victor Sanh, Julien Chaumond, and Clement Delangue. 2019.
\newblock \href {http://arxiv.org/abs/1901.08149} {Transfertransfo: {A}
  transfer learning approach for neural network based conversational agents}.
\newblock \emph{CoRR}, abs/1901.08149.

\bibitem[{Xie et~al.(2019)Xie, Dai, Hovy, Luong, and
  Le}]{DBLP:journals/corr/abs-1904-12848}
Qizhe Xie, Zihang Dai, Eduard~H. Hovy, Minh{-}Thang Luong, and Quoc~V. Le.
  2019.
\newblock \href {http://arxiv.org/abs/1904.12848} {Unsupervised data
  augmentation}.
\newblock \emph{CoRR}, abs/1904.12848.

\bibitem[{Xu et~al.(2020)Xu, Li, Yang, Ren, Ren, Chen, and
  Ma}]{DBLP:journals/corr/abs-2002-02153}
Minghong Xu, Piji Li, Haoran Yang, Pengjie Ren, Zhaochun Ren, Zhumin Chen, and
  Jun Ma. 2020.
\newblock \href {http://arxiv.org/abs/2002.02153} {A neural topical expansion
  framework for unstructured persona-oriented dialogue generation}.
\newblock \emph{CoRR}, abs/2002.02153.

\bibitem[{Zeng et~al.(2019)Zeng, Zhang, Xiang, Wang, and
  Ji}]{DBLP:journals/access/ZengZXWJ19}
Daojian Zeng, Haoran Zhang, Lingyun Xiang, Jin Wang, and Guoliang Ji. 2019.
\newblock \href {https://doi.org/10.1109/ACCESS.2019.2923057} {User-oriented
  paraphrase generation with keywords controlled network}.
\newblock \emph{{IEEE} Access}, 7.

\bibitem[{Zhang et~al.(2018)Zhang, Dinan, Urbanek, Szlam, Kiela, and
  Weston}]{DBLP:conf/acl/KielaWZDUS18}
Saizheng Zhang, Emily Dinan, Jack Urbanek, Arthur Szlam, Douwe Kiela, and Jason
  Weston. 2018.
\newblock \href {https://doi.org/10.18653/v1/P18-1205} {Personalizing dialogue
  agents: {I} have a dog, do you have pets too?}
\newblock In \emph{ACL}.

\bibitem[{Zhang et~al.(2020)Zhang, Kishore, Wu, Weinberger, and
  Artzi}]{DBLP:conf/iclr/ZhangKWWA20}
Tianyi Zhang, Varsha Kishore, Felix Wu, Kilian~Q. Weinberger, and Yoav Artzi.
  2020.
\newblock \href {https://openreview.net/forum?id=SkeHuCVFDr} {Bertscore:
  Evaluating text generation with {BERT}}.
\newblock In \emph{ICLR}.

\bibitem[{Zhao et~al.(2017)Zhao, Zhao, and
  Esk{\'{e}}nazi}]{DBLP:conf/acl/ZhaoZE17}
Tiancheng Zhao, Ran Zhao, and Maxine Esk{\'{e}}nazi. 2017.
\newblock \href {https://doi.org/10.18653/v1/P17-1061} {Learning
  discourse-level diversity for neural dialog models using conditional
  variational autoencoders}.
\newblock In \emph{ACL}.

\bibitem[{Zhao et~al.(2011)Zhao, Hachiya, Niu, and
  Sugiyama}]{DBLP:conf/nips/ZhaoHNS11}
Tingting Zhao, Hirotaka Hachiya, Gang Niu, and Masashi Sugiyama. 2011.
\newblock \href
  {http://papers.nips.cc/paper/4264-analysis-and-improvement-of-policy-gradient-estimation}
  {Analysis and improvement of policy gradient estimation}.
\newblock In \emph{NIPS}.

\bibitem[{Zhou et~al.(2018)Zhou, Young, Huang, Zhao, Xu, and
  Zhu}]{DBLP:conf/ijcai/ZhouYHZXZ18}
Hao Zhou, Tom Young, Minlie Huang, Haizhou Zhao, Jingfang Xu, and Xiaoyan Zhu.
  2018.
\newblock \href {https://doi.org/10.24963/ijcai.2018/643} {Commonsense
  knowledge aware conversation generation with graph attention}.
\newblock In \emph{IJCAI}.

\end{thebibliography}
\bibliographystyle{acl_natbib}

\appendix

\section{Implementation Details}
We obtain the \personachat~dataset from ParlAI repository\footnote{\url{http://parl.ai/downloads/personachat/personachat.tgz}}. For COMET expansions, we use the code\footnote{\url{https://github.com/atcbosselut/comet-commonsense}} released by the authors of COMET \cite{DBLP:conf/acl/BosselutRSMCC19}.
We performed BPE tokenization with the  GPT2Tokenizer\footnote{\url{https://huggingface.co/transformers/model_doc/gpt2.html}}.
\paragraph{Network architectures}

For the generator network, we use GPT2 (Transformer with 12 layers, 768 hidden size, 12 heads--- \texttt{gpt2-small}\footnote{\url{https://github.com/huggingface/transfer-learning-conv-ai}}) following the state-of-the-art model \cite{DBLP:journals/corr/abs-1901-08149} from Conv-AI2 competition.
\citet{DBLP:journals/corr/abs-1901-08149} also leveraged incorrect responses given a dialog history from \personachat~as negative samples in an auxiliary loss to encourage the correct candidate to obtain the highest likelihood compared to the incorrect ones. However, we did not find any improvement using this loss in \compac.
\compac~has total of 164 Million parameters whereas GPT2 based baseline has 124 Million parameters.

\paragraph{Hyperparameters}
Following \cite{DBLP:journals/corr/abs-1901-08149} we use history size 2 (i.e.~4 previous utterances).
We use AdamW optimizer \cite{DBLP:journals/corr/abs-1711-05101} and the learning rate was set at $6.25e-5$ with a linear decay of step size $10^{-1}$ per epoch. The baseline in REINFORCE was done with a discounted moving average with a ratio of 0.99. The REINFORCE loss coefficient was set at 0.8 and the language modeling loss coefficient was set to 1.0.

\paragraph{Training}
Each model converged in 3 epochs on an average with batch size 4 in a TITAN X (Pascal) GPU that took 12 hours in total. While training, we only observe perplexity on the validation set to employ an early-stopping criteria.

\section{Evaluation}

\noindent \textbf{Automatic Evaluation}
During dialog quality evaluation, perplexity is measured by adapting the official evaluation protocol from the Conv-AI2 challenge\footnote{\url{https://github.com/facebookresearch/ParlAI/blob/master/projects/convai2}}.

To assess persona grounding, we use Dialogue Natural Language Inference (DNLI) dataset \cite{DBLP:conf/acl/WelleckWSC19} that has persona-utterances pairs under three classes---entailment, neutral, and contradiction. We gather all the entailment pairs including all splits that resulted in 44,000 persona-utterance pairs.
Then we map with the \personachat~test set to obtain 4,613 utterances associated with a ground truth persona. 

For assessing conditional generation performance, we use BERT score from the publicly available repository\footnote{\url{https://github.com/Tiiiger/bert_score}}.

\begin{figure*}[t]
    \centering
    \includegraphics[width=\linewidth]{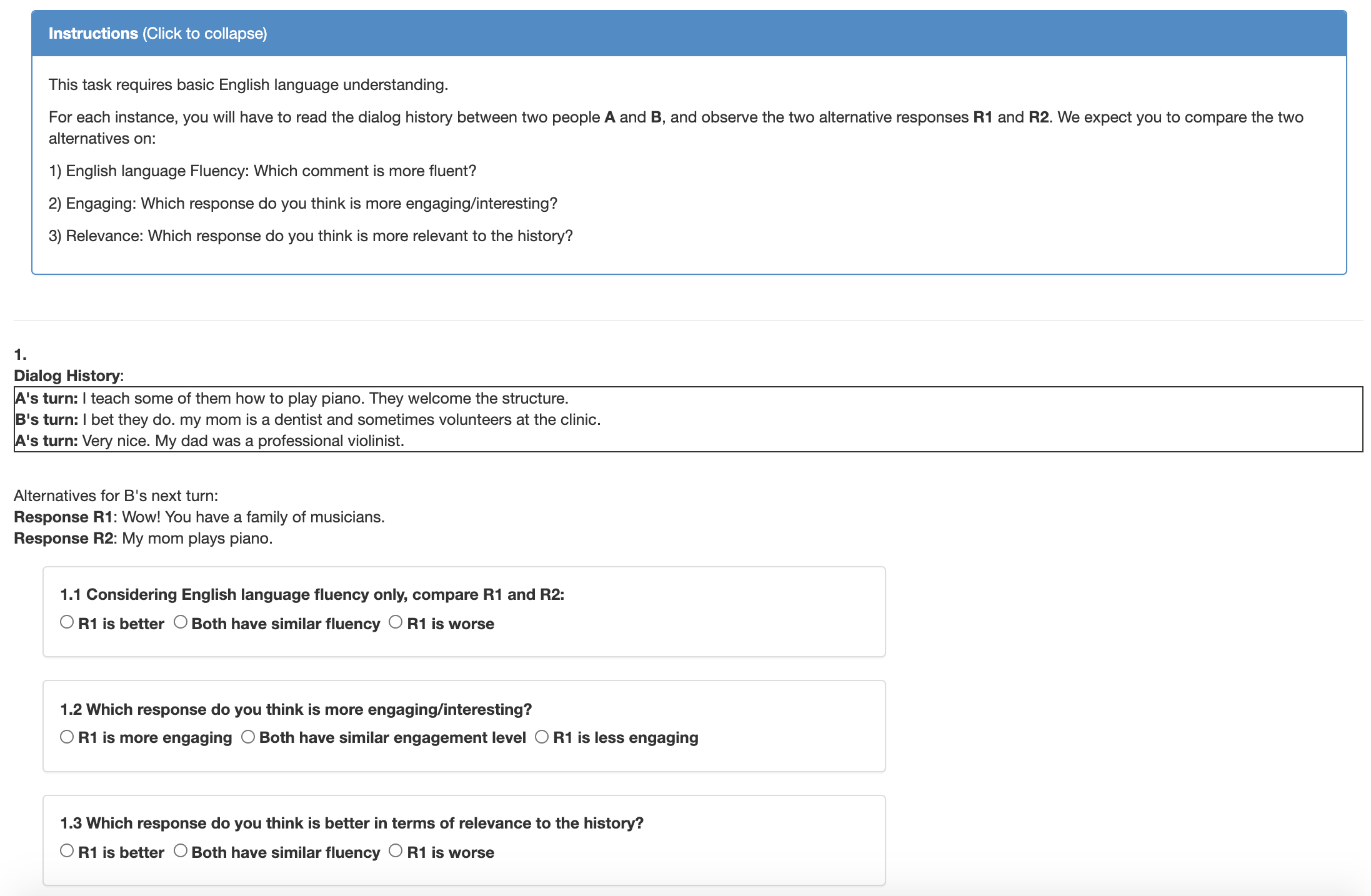}
    \caption{Human evaluation for pairwise comparison between \compac~and another baseline.}
    \label{fig:human_eval1}
\end{figure*}

\noindent \textbf{Human Evaluation}
For human evaluation, we hired two Anglophone (Lifetime HIT acceptance \% $>$ 80) annotators for every human-evaluated test generation. \Cref{fig:human_eval1} shows a sample question for a human judge for the pairwise comparison of a response generated by \compac~and a response generated by a baseline for three aspects---fluency, engagement, and coherence.

\begin{figure*}[t]
    \centering
    \includegraphics[width=\linewidth]{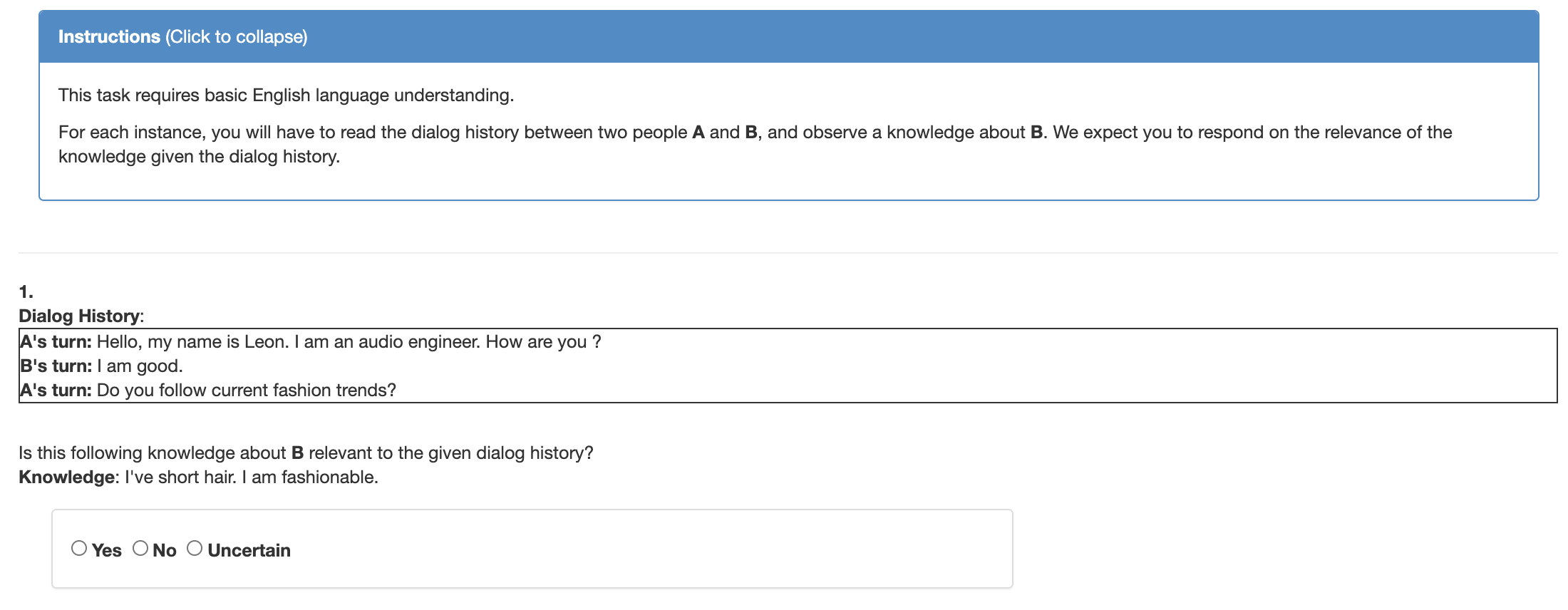}
    \caption{Human evaluation for assessment of persona grounding performance with expansions.}
    \label{fig:human_eval2}
\end{figure*}

While measuring persona grounding, we used a similar setup where we provided a dialog history and a sampled expansion and asked `Is this knowledge relevant to the given dialog history?'--- with three options ---`Yes', `No', and `Uncertain' (See \Cref{fig:human_eval2}). Similar to the previous human evaluation study, we hired two Anglophone (Lifetime HIT acceptance \% $>$ 80) annotators for each question. We find the inter-annotator agreement, as measured by
Cohen's
kappa was 0.76.

\section{Generation Examples}
\Cref{tab:sample2,tab:sample3,tab:sample4,tab:sample5} present generations from \compac~for sampled dialog histories with comparison across baselines.

\begin{table}[t]
\small
\centering
\begin{tabular}{@{}l@{}}
\toprule
\begin{tabular}{@{}l@{}}
\textbf{Persona:} \\
I've short hair.\\
I have a pet dog.\\
I like to go hiking.\\ \midrule
\midrule
\textbf{User:} Do you follow current fashion trends?\\ \midrule
\textbf{GPT2} \shortcite{DBLP:journals/corr/abs-1901-08149}\textbf{:}
Yes. What about you?\\ \midrule
\textbf{\compac-original:}
I have a pet dog.\\ \midrule
\textbf{\compac-revised:}
I think you really need to go shopping.\\ \midrule
\textbf{\compac-paraphrase:}
I like shopping. \\ \midrule
\textbf{\compac:}
Yes, I prefer to be fashionable.\\ \bottomrule
\end{tabular}
\end{tabular}
\caption{Sample generations by different models.}
\label{tab:sample2}
\end{table}

\begin{table}[t]
\small
\centering
\begin{tabular}{@{}l@{}}
\toprule
\begin{tabular}{@{}l@{}}
\textbf{Persona:} \\
I just want to have fun with my friends.\\
I don't drink or do drugs or anything.\\
I am 19 and I cannot wait to move out of my parents home.
\\ \midrule
\midrule
\textbf{User:} Are you enjoying life ?\\ \midrule
\textbf{GPT2} \shortcite{DBLP:journals/corr/abs-1901-08149}\textbf{:}
Yes! What about you? \\ \midrule
\textbf{\compac-original:}
Hey yes! I don't drink. \\ \midrule
\textbf{\compac-revised:}
Yes. I like to go out with friends.\\ \midrule
\textbf{\compac-paraphrase:}
I don't want to be alone. \\ \midrule
\textbf{\compac:}
Not really. I want to have a party.\\ \bottomrule
\end{tabular}
\end{tabular}
\caption{Sample generations by different models.}
\label{tab:sample3}
\end{table}

\begin{table}[t]
\small
\centering
\begin{tabular}{@{}l@{}}
\toprule
\begin{tabular}{@{}l@{}}
\textbf{Persona:} \\
I've rainbow hair.\\
I am an animal activist.\\
I spend my time bird watching with my cats.
\\ \midrule
\midrule
\textbf{User:} What motivates you to work with animals?\\ \midrule
\textbf{GPT2} \shortcite{DBLP:journals/corr/abs-1901-08149}\textbf{:}
I love bird-watching. \\ \midrule
\textbf{\compac-original:}
I love bird-watching. What about you? \\ \midrule
\textbf{\compac-revised:}
I love animals. \\ \midrule
\textbf{\compac-paraphrase:}
I am in love with animals.\\ \midrule
\textbf{\compac:}
I want to protect animals to make a\\difference in this world.\\ \bottomrule
\end{tabular}
\end{tabular}
\caption{Sample generations by different models.}
\label{tab:sample4}
\end{table}

\begin{table}[t]
\small
\centering
\begin{tabular}{@{}l@{}}
\toprule
\begin{tabular}{@{}l@{}}
\textbf{Persona:} \\
I'm always wearing a hat.\\
I've a broken arm.  \\
My friends are all skateboarders.\\ \midrule
\midrule
\textbf{User:} Do you skate? I do.\\ \midrule
\textbf{GPT2} \shortcite{DBLP:journals/corr/abs-1901-08149}\textbf{:}
Yes. How about you? \\ \midrule
\textbf{\compac-original:}
I wear a hat.\\ \midrule
\textbf{\compac-revised:}
Yes. My friends are skateboarders. \\ \midrule
\textbf{\compac-paraphrase:}
That's great. My friends are\\skateboarders. \\ \midrule
\textbf{\compac:}
My friends and I go to the park for skateboarding.\\ \bottomrule
\end{tabular}
\end{tabular}
\caption{Sample generations by different models.}
\label{tab:sample5}
\end{table}





\end{document}